\documentclass[letterpaper, 10 pt, conference]{ieeeconf}  

\IEEEoverridecommandlockouts                              
\usepackage{amsmath,bm}    			
\usepackage{amssymb}
\usepackage{amsfonts}
\usepackage{graphicx}  				

\usepackage[croatian]{babel}  
\usepackage[utf8]{inputenc}  	
\usepackage[T1]{fontenc}
\usepackage{ae,aecompl}     	

\usepackage{microtype}				

\usepackage{subfig}

\usepackage{tabularx}
\usepackage{booktabs}
\usepackage{enumerate}				
\usepackage{algorithm2e}			
\usepackage{todonotes}				
\usepackage{dirtree}					
\usepackage{hyperref}					

\usepackage{graphicx}

\usepackage{lmodern}
\usepackage{nccmath}
\usepackage{scalerel,stackengine}
\stackMath
\newcommand\reallywidehat[1]{%
	\savestack{\tmpbox}{\stretchto{%
			\scaleto{%
				\scalerel*[\widthof{\ensuremath{#1}}]{\kern.1pt\mathchar"0362\kern.1pt}%
				{\rule{0ex}{\textheight}}
			}{\textheight}%
		}{2.4ex}}%
	\stackon[-6.9pt]{#1}{\tmpbox}%
}
\usepackage{calrsfs}
\DeclareMathAlphabet{\pazocal}{OMS}{zplm}{m}{n}

\graphicspath{{./figures/}}
\usepackage{float}

\title{\LARGE \bf
Exploiting Null Space in Aerial Manipulation through Model-In-The-Loop Motion Planning
}
\author{Antun Ivanovic, Marko Car, Matko Orsag, Stjepan Bogdan
	\thanks{Authors are with Faculty of Electrical and Computer Engineering,
        University of Zagreb, 10000 Zagreb, Croatia
        {\tt\small (antun.ivanovic, marko.car, matko.orsag, stjepan.bogdan) at fer.hr}}}%

\makeatletter
\newcommand{\removelatexerror}{\let\@latex@error\@gobble}

\makeatother

\begin{document}
\maketitle

\thispagestyle{empty}
\pagestyle{empty}

\begin{abstract}

This paper presents a method for aerial manipulator end-effector trajectory tracking by encompassing dynamics of the Unmanned Aerial Vehicle (UAV) and null space of the manipulator attached to it in the motion planning procedure. The proposed method runs in phases. Trajectory planning starts by not accounting for roll and pitch angles of the underactuated UAV system. Next, we propose simulating the dynamics on such a trajectory and obtaining UAV attitude through the model. The full aerial manipulator state obtained in such a manner is further utilized to account for discrepancies in planned and simulated end-effector states. Finally, the end-effector pose is corrected through the null space of the manipulator to match the desired end-effector pose obtained in trajectory planning. Furthermore, we have applied the TOPP-RA approach on the UAV by invoking the differential flatness principle. Finally, we conducted experimental tests to verify effectiveness of the planning framework.

\end{abstract}
\section{Introduction}
The field of aerial robotics is nowadays gaining ever more interest. Still, the most popular vehicles are underactuated Vertical Takeoff and Land (VTOL) multirotor UAVs since they do not require specialized infrastructure to become airborne. Every day new applications are explored such as wind turbine inspection, geodetic mapping, all the way towards filming industry. Increase in supply dropped the prices of off-the-shelf UAVs, making multirotors affordable to both professionals and general public. Nevertheless, the research community is still struggling to unlock the full potential of multirotors. One of the most promising fields for future applications is aerial manipulation, where multirotor UAVs are endowed with various manipulators. This gives such vehicles the ability to interact and change the environment rather than just observe it and collect information. In \cite{korpela2014} researchers envisioned insertion tasks for aerial manipulators. In line with that, researchers in \cite{steich2016} used camera equipped aerial manipulator to inspect tree cavities. Authors of \cite{tsukagoshi2015} proposed door opening task using an aerial manipulator, while researchers in \cite{suarez2017} designed a human-size dual-arm lightweight manipulator. Another research topic worth mentioning is autonomous delivery tasks \cite{arbanas2018}, \cite{kuru2019}. As evidenced by this prior work, interaction with the environment universally boils down to careful motion planning and its execution. In this paper a UAV equipped with a 3 degree of freedom manipulator is deployed on inspection missions in a constrained environment. The idea behind the concept is to exploit the null space of the aerial manipulator in order to inspect places which would be impossible to reach with a standard UAV.

The inspiration for this work stems from project \textit{Specularia}. The main task of the project is to develop an autonomous robotic system for plant inspection and treatment in a structured greenhouse. Aerial manipulator is deployed to provide closeup images of plants with camera mounted at the end-effector avoiding harming the plants at the same time. Therefore, the end-effector must execute motion close to fragile stems and leaves without damaging them. This kind of motion can be approximated with peg-in-hole insertion task where the end-effector follows a certain trajectory and compensates for dynamic movement of the UAV through its null space motion. Furthermore, such a system can be deployed on infrastructure inspection to insert the camera in crevices and other hard-to-reach places. 

\begin{figure}[t]
	\centering
	\includegraphics[width=\columnwidth]{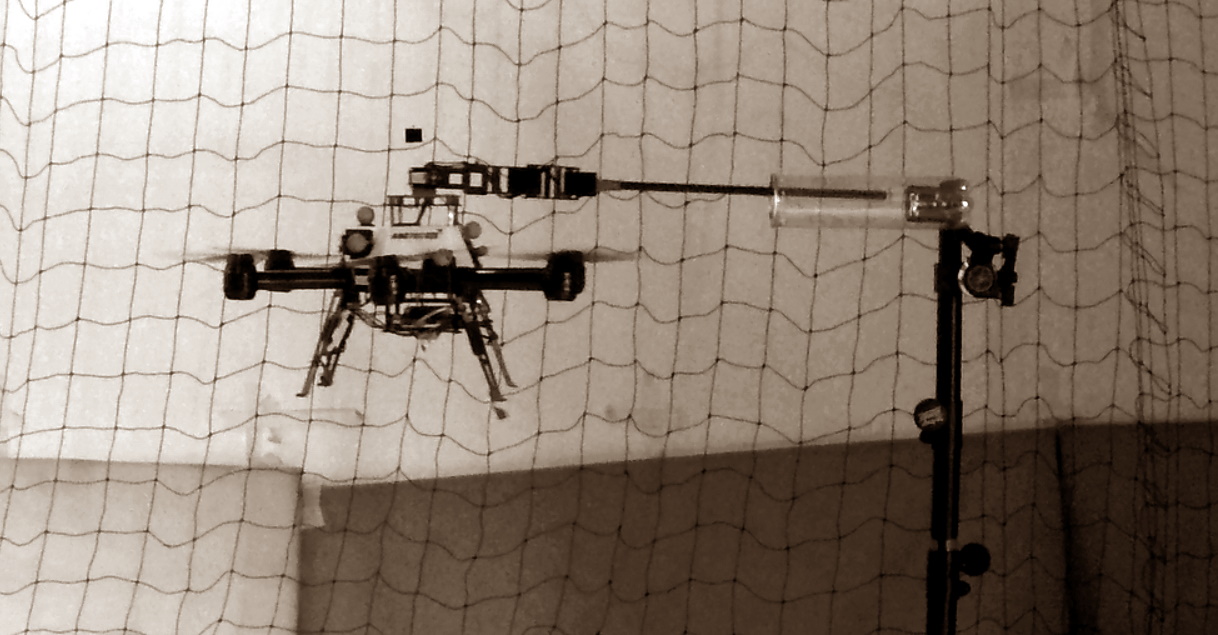}
	\caption{AscTec NEO equipped with 3DOF manipulator performing peg-in-hole insertion task. An empty cylinder acts as a mock-up version of the task constrained space. Image is taken during experimental verification of the proposed trajectory planning algorithm. }
	\label{fig:neo_peg_in_hole}
	\vspace{-0.55cm}
\end{figure}

Motion planning is a very well established topic in robotics. One of the earliest path planners able to cope with high dimensional unknown environments was Rapidly-exploring Random Tree (RRT) published in \cite{lavlle2000}. In this paper we utilize such a planner as a first attempt to find a feasible path in a constrained environment. We begin by introducing task space constraints, inspired by researchers in \cite{stilman2010} which were able to utilize null space of the manipulator to execute constrained manipulator motions in cluttered environment. Similarly, the researchers in \cite{cefalo2015} address the motion planning problem for underactuated robot in task space. In this work we use task space to describe additional constraints aerial manipulator faces in order to complete the task, at the same time maintaining safe distance from the fragile environment.

Building on top of the planned path, trajectory planners such as CHOMP \cite{chomp} and STOMP \cite{stomp}, ensure smooth motion execution. In our work we apply the Time Optimal Path Parametrization (TOPP) \cite{topp}, which is a numerical approach that provides very fast trajectory generation. It is based on a "bang-bang" principle on generalized joint torques which yields a time optimal trajectory. An extension of this algorithm, which offers even greater success rate, called TOPP by Reachability Analysis (TOPP-RA), proposed in \cite{toppra} enables us to plan feasible trajectories in real time. 

Motion planning for aerial manipulators has been largely explored for single vehicle \cite{yang2014} and multiple vehicles \cite{fink2011}. It relies on the principle of differential flatness, which enables engineers to calculate control inputs directly from the planned motion. Important results from \cite{yuksel2016differential} show that differential flatness exists for rotorcraft UAVs endowed with manipulators placed in the vehicle's center of mass. Most approaches for UAV trajectory planning tend to use convex optimization to minimize some criterion. Researchers in \cite{mellinger2011} proposed an algorithm that minimizes $4^{th}$ derivative of position. Authors of \cite{Richter2016} extend that work by improving numerical stability of the approach. The work presented in \cite{bo2017} deals with trajectory planning for multiple UAVs. A more recent research has started looking into considering dynamics of the aerial manipulator in motion planning \cite{tognon2018}. In the proposed work, we extend the current body of work by exploring how the manipulator null space can be exploited to account for the explored dynamics of the aerial vehicle. Even though it incorporates dynamics in the planning procedure, the proposed algorithm can be executed in real time, unlike well known and rather slow kinodynamic approaches \cite{lavalle2001randomized}.

The main contribution of this paper is the proposed planning framework that utilizes the dynamic model of the system to calculate the necessary compensation via manipulator null space in order to keep the end-effector pose on the desired trajectory, avoiding obstacles at the same time. This kind of approach is feasible since we decouple the UAV and the manipulator configuration space (C-space)  during the compensation phase. Furthermore, we have applied the TOPP-RA approach on the UAV by invoking the differential flatness principle. Finally, we conducted experimental tests to verify effectiveness of the planning framework.

The paper is organized as follows. In Section \ref{sec:model} we derive the kinematic and dynamic model of the aerial manipulator. Next, in Section \ref{sec:planning} the proposed motion planning framework is presented. This is followed by experimental verification on a canonical peg-in-hole insertion task in Section \ref{sec:experimental}. Finally, the conclusion is given in Section \ref{sec:conclusion}.

\section{Mathematical model} \label{sec:model}

\begin{figure}[t]
\centering
\includegraphics[width=0.95\columnwidth]{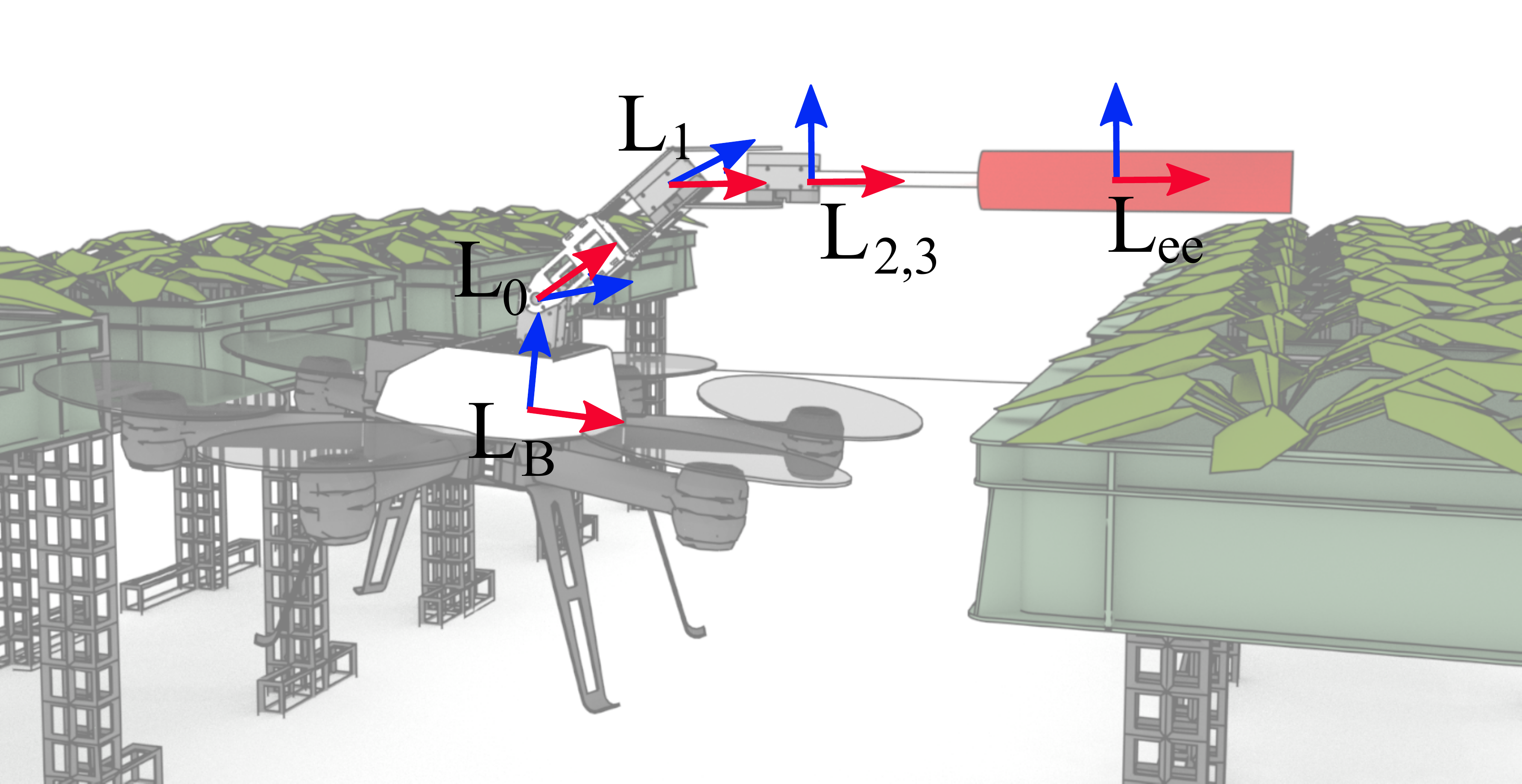}
\caption{Coordinate frames of the UAV and the single-arm manipulator. The origins of $L_3$ and $L_2$ coincide since $L_3$ represents a virtual joint required to obtain manipulator's DH parameters. Such an aerial manipulator faces multiple constraints, for instance the body of the UAV must remain below the plants while the end effector must reach the task constraints depicted with a virtual red cylinder.}
\label{fig:single_arm_manipulator_frames}
\vspace{-0.45cm}
\end{figure}

In this section we present the generalized mathematical model of an aerial manipulator. Fig. \ref{fig:single_arm_manipulator_frames} depicts an aerial vehicle with a manipulator attached to its body.

\subsection{Kinematics}
The inertial frame is defined as $L_\mathit{W}$. The gravity vector is directed along negative $z_\mathit{W}$ axis of the inertial frame. The body-fixed frame $L_\mathit{B}$ is attached to the center of gravity of the UAV body. We can define the generalized coordinates of the UAV as  $\mathbf{q}_B = \begin{bmatrix} \mathbf{p}_B^T & \mathbf{\Theta}_B^T \end{bmatrix}^T \in \mathbb{R}^6$, where $\mathbf{p}_B = \begin{bmatrix}
x & y & z
\end{bmatrix}^T$ represents the position of the UAV in the inertial frame and $\mathbf{\Theta}_B = \begin{bmatrix}
\phi & \theta & \psi
\end{bmatrix}^T$ represents the attitude vector. 

The $M$ Degree-of-Freedom (DoF) manipulator is considered to be rigidly attached to the body of the UAV with its origin in the frame $L_0$, which also coincides with the first joint of the manipulator. The kinematic chain of the manipulator is represented with frames $(L_0, \dots, L_{M-1}, L_{ee})$, where the latter denotes the end-effector frame. The full kinematic chain of the aerial manipulator can be written as:
\begin{equation} \label{eqn:kinematic_chain}
	T_\mathit{W}^{ee} = T_\mathit{W}^\mathit{B} \cdot T_\mathit{B}^{L_0} \cdot T_{L_0}^{ee},
\end{equation}
\noindent where $T_a^b \in \mathbb{R}^{4 \times 4}$ defines a homogeneous transformation matrix between the coordinate frames $a$ and $b$. $T_\mathit{B}^{L_0}$ is fixed a transformation defined by the manipulator placement on the UAV body, and $T_{L_0}^{ee}$ can be obtained through direct kinematics of the manipulator.

\subsection{Dynamics} \label{sec:dynamics}
To cope with dynamic effects during in-flight motion, we derive a full dynamical model of the aerial manipulator. We previously defined $\mathbf{q}_B$ as the general coordinates of the UAV. The linear velocity $\dot{\mathbf{p}}_B$ is a time derivative of the position in the world frame. The angular velocity in the world frame is $\bm{\omega}_B^T = p\bm{x}_B + q\bm{y}_B + r\bm{z}_B$, where $p$, $q$ and $r$ are angular velocities in the body frame which can be mapped to the world frame. This yields the velocity of the generalized coordinates as  $\dot{\mathbf{q}}_B = \begin{bmatrix}
\dot{\mathbf{p}}_B^T & \bm{\omega}_B^T
\end{bmatrix}^T \in \mathbb{R}^6$.
The propulsion system of the UAV consists of multiple propellers rigidly attached to the body of the UAV. Number of the propellers $n_p$ may vary depending on the configuration of the UAV. Each propeller is considered to produce thrust force along $z_\mathit{B}$ axis, which makes the UAV underactuated system. We can express the force vector as $\mathbf{f} = \begin{bmatrix}
f_1 & \dots & f_{n_p} 
\end{bmatrix}^T \in \mathbb{R}^{n_p}$. 

The generalized coordinates of the manipulator are defined as $\mathbf{q}_M = \begin{bmatrix}
q_1 & \dots & q_M
\end{bmatrix}^T \in \mathbb{R}^M$, where each joint can be either rotational or prismatic. In both cases, the Denavit-Hartenberg convention suggests that $z$ axis of each joint represents the axis of actuation: rotation for rotational joints, and translation for prismatic joints. The velocity of manipulator joints is a time derivative of their position $\dot{\mathbf{q}}_M = d \mathbf{q}_M/dt \in \mathbb{R}^M$. Control input for each joint is considered to be generalized torque along its respective axis. In this case, the vector containing all generalized torques is written as $\bm{\tau} = \begin{bmatrix}
\tau_1 & \dots & \tau_M
\end{bmatrix}^T \in \mathbb{R}^M $.

The generalized coordinates of the aerial manipulator can be expressed as $\mathbf{q} = \begin{bmatrix}
\mathbf{q}_B^T & \mathbf{q}_M^T
\end{bmatrix}^T \in \mathbb{R}^{6+M}$. Velocity can be obtained in the same manner: $\dot{\mathbf{q}} = \begin{bmatrix}
\dot{\mathbf{q}}_B^T & \dot{\mathbf{q}}_M^T
\end{bmatrix}^T \in \mathbb{R}^{6+M}$. The control input vector of the whole system can be written as: $\mathbf{u}^* = \begin{bmatrix}
\mathbf{f}^T & \bm{\tau}^T
\end{bmatrix}^T \in \mathbb{R}^{n_p+M}$. Although $\bm{u}$ captures all the control inputs, it is more suitable to write generalized input vector due to underactuated properties of the multirotor UAVs. Due to underactuated properties of the multirotor UAVs it is necessary to define $\mathbf{u}_\textsc{uav} = \begin{bmatrix} u_1 & u_2 & u_3 & u_4 \end{bmatrix}^T$, where $u_1$ represents the net thrust and $u_2$, $u_3$ and $u_4$ are moments around body frame axes. Depending on the configuration of the multirotor UAV, forces produced by each rotor are $\bm{f} = \bm{K}^{-1} \cdot \mathbf{u}_\textsc{uav}$, where matrix $\bm{K} \in \mathbb{R}^{4 \times n_p}$ relates the propeller forces and control signals. The full system's dynamics can now be captured as:
\begin{equation} \label{eqn:newton_euler_dynamics}
	M(\mathbf{q})\ddot{\mathbf{q}} + \mathbf{c}(\mathbf{q}, \dot{\mathbf{q}}) + \mathbf{g}(\mathbf{q}) = \mathbf{u},
\end{equation}
\noindent where $M(\mathbf{q}) \in \mathbb{R}^{(6+M)\times(4+M)}$ is inertia matrix, $\mathbf{c}(\mathbf{q}, \dot{\mathbf{q}}) \in \mathbb{R}^{4+M}$ is the vector of centrifugal and Coriolis forces, $\mathbf{g}(\mathbf{q}) \in \mathbb{R}^{4+M}$ is the gravitational term, and $\mathbf{u} = \begin{bmatrix} \mathbf{u}_\textsc{uav}^T & \bm{\tau}^T \end{bmatrix}^T$ is the generalized control input. The state of the robot at any given point in time is defined with position and velocity: $\mathbf{x} = \begin{bmatrix}
\mathbf{q}^T & \dot{\mathbf{q}}^T 
\end{bmatrix}^T \in \mathbb{R}^{2(6+M)}$.



\section{Planning framework} \label{sec:planning}

Most common manipulation problems consider tracking a desired trajectory or a series of waypoints with the end-effector. This is often the goal in inspection, grinding, insertion and many other tasks. In case of an aerial manipulator, the same problem can be solved by encompassing the UAV dynamics in the planning procedure.

\begin{figure}[t]
	\centering
	\includegraphics[width=\columnwidth]{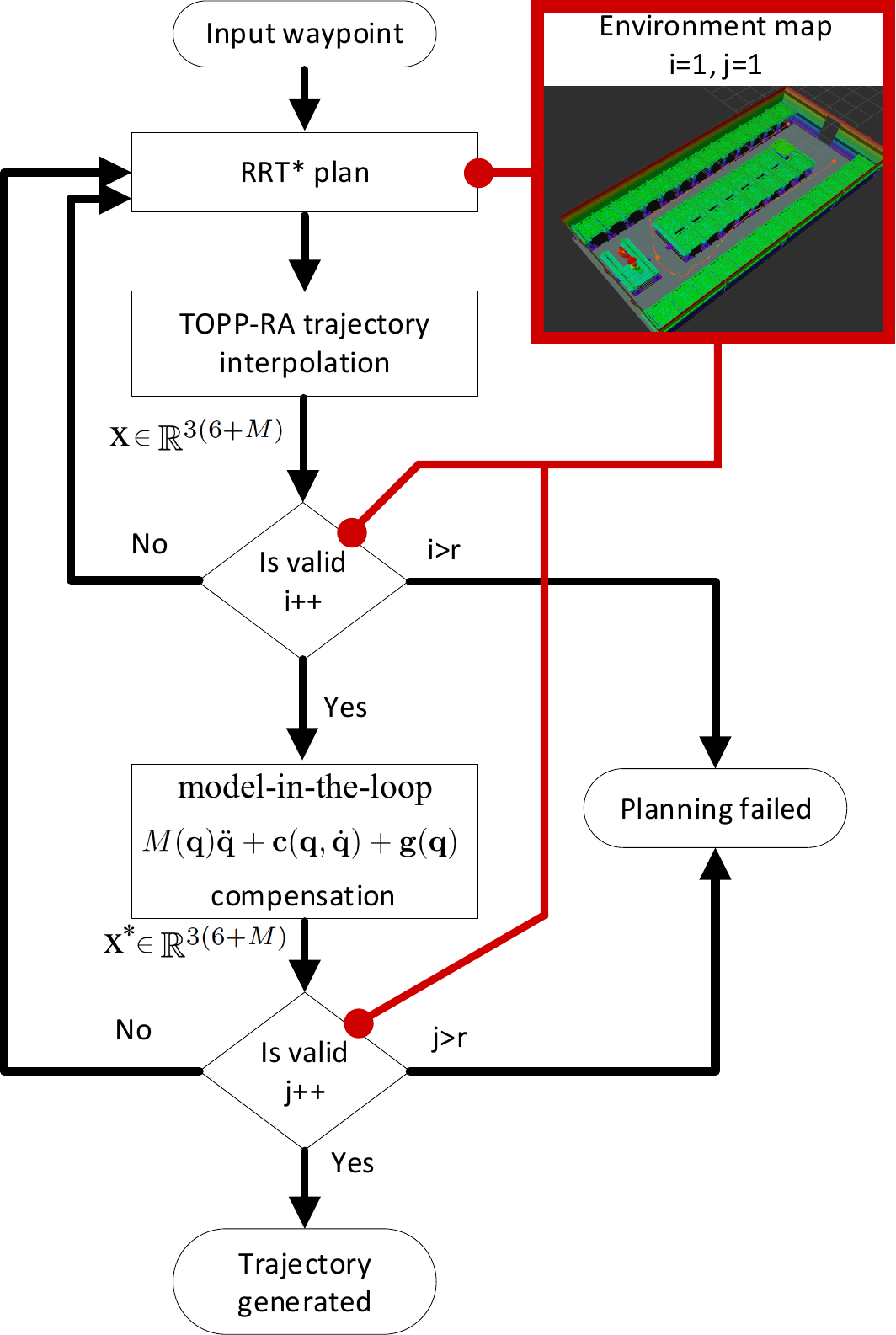}
	\caption{Trajectory planning algorithm verifies aerial manipulator pose in three stages, starting with simplified model RRT\* search, which is then tested using TOPP-RA trajectory interpolation. The complete model of the aerial manipulator is then used to calculate trajectory compensation  by including system dynamics.}
	\label{fig:planning_algorithm}
	\vspace{-0.45cm}
\end{figure}

The planning framework in this paper consists of several steps to account for the aforementioned dynamical properties of the aerial vehicle and to ensure the final trajectory is feasible and collision free. Fig. \ref{fig:planning_algorithm} illustrates the algorithm used to obtain such trajectories. Given a series of waypoints and the map of the environment, the planner outputs a trajectory for the given task. Note that this planning framework is not limited to aerial manipulators, by using different dynamic models it can be adapted to all types of vehicles i.e. ground or marine robots.

There are three main phases in the proposed planning framework: path planning in a high dimensional configuration space of the aerial manipulator; trajectory planning based on the obtained path; and utilizing the manipulator null space to compensate the end-effector pose for unknown dynamical effects through simulation.

\subsection{Path planning}
Within this paper, the path planning procedure is done in the C-space of the robot. The main task of path planner is to find obstacle-free piecewise straight line path in high dimensional robot C-space. For our purposes, we use the well known RRT* planner, but other planners such as Probabilistic Road Map(PRM), A* or D* can be used as well. The main advantage of such sampling based planners is the very fast exploration of an unknown high-dimensional space, which is possible due to increase in computing power. Nevertheless, when executing fast trajectories, one must be careful and aware of the trajectory deviation from piecewise straight line that was planned. The path planner also inherently assumes the attitude of the UAV will not change, however, due to underactuated nature of the multirotor UAV in use, the attitude changes. As the UAV rotates while executing a motion the manipulator's end-effector pose will not track the desired trajectory. This behaviour is addressed by exploiting the null-space of the manipulator to compensate the desired trajectory tracking error in Section \ref{sec:end_effector_pose_comp}.

To account for the underactuated nature of the UAV, using the same notation as in Section \ref{sec:model}, we introduce control space generalized coordinates vector for the UAV as $\mathbf{q}_B^* = \begin{bmatrix} \mathbf{p}_B^T & \psi \end{bmatrix}^T \in \mathbb{R}^{4}$. This yields the control space aerial manipulator generalized coordinates vector $\mathbf{q}^* = \begin{bmatrix} (\mathbf{q}_B^*)^T & \mathbf{q}_M^T \end{bmatrix}^T \in \mathbb{R}^{4+M}$. The planning space for the RRT* algorithm is spanned with generalized coordinates $\mathbf{q}^*$ and the desired waypoints are expressed in terms of the control space generalized coordinates.


The input to the path planner is a set of $m \geq 2$ waypoints. The simplest example would be providing only two waypoints, i.e. start and goal. However, in general case there could be multiple points of interest that have to be visited. The task of the path planner is to find an obstacle-free piecewise straight path between user provided waypoints, resulting in a set of point:
\begin{equation} \label{eqn:path}
	\mathcal{P} = \left\{\mathbf{p}_i \mid \mathbf{p}_i \in \mathbb{R}^{4+M}, i \in (0, 1, \dots, n) \right\},
\end{equation}
\noindent where $n \geq 2$ is the number of points in the planned path and $\mathbf{p}_i$ are the control space generalized coordinates. One can observe that if there are no obstacles between the provided waypoints, the path planning algorithm simply returns provided waypoints. On the other hand, should there be obstacles, the path planning algorithm returns piecewise straight line path with additional points for avoiding obstacles.

\subsection{Trajectory planning}
The trajectory planning introduces dynamic constraints, namely maximum velocity and acceleration, in the planning framework. According to Pham \cite{topp}, there are three families of trajectory planning methods: dynamic programming; convex optimization; and numerical integration. The first two methods plan for the optimal trajectory according to the specified criteria. While providing the optimal solution, the main disadvantage of these methods is a long runtime while searching for the solution. On the other hand, the TOPP approach provides the time optimal solution using the "bang-bang" principle on generalized torques of the actuators and runs much faster compared to dynamic programming and convex optimization. In this paper, we are using an extension of the TOPP approach, called TOPP-RA, proposed in \cite{toppra}.

Similar to path planning, the trajectory is also planned in the control space generalized coordinates. Therefore, the velocity can be written as $\dot{\mathbf{q}}^* = \begin{bmatrix} (\dot{\mathbf{q}}_B^*)^T & \dot{\mathbf{q}}_M^T \end{bmatrix}^T \in \mathbb{R}^{4+M}$ and acceleration is defined as $\ddot{\mathbf{q}}^* = \begin{bmatrix} (\ddot{\mathbf{q}}_B^*)^T & \ddot{\mathbf{q}}_M^T \end{bmatrix}^T \in \mathbb{R}^{4+M}$. Including velocity and acceleration, a single trajectory point in time can be written as $\mathbf{w}^* = \begin{bmatrix} (\mathbf{q}^*)^T & (\dot{\mathbf{q}}^*)^T & (\ddot{\mathbf{q}}^*)^T \end{bmatrix}^T \in \mathbb{R}^{3(4+M)}$. Note that we use such notation for $\mathbf{w}^*$ simply because it is composed of control space generalized coordinates.

The input to the trajectory planning algorithm is path, a series of points defined with equation \eqref{eqn:path}. TOPP-RA also requires a set of dynamic constraints that correspond to the maximum velocity and acceleration of each joint. As shown in \cite{mellinger2011}, the multirotor UAV is a differentially flat system. Such systems have a unique property that all states and inputs can be expressed in terms of outputs and outputs’ derivatives. If states and inputs can be written in terms of the generated trajectory, it is possible to calculate inputs required to track that trajectory and check if the trajectory satisfies feasibility conditions. Given reasonable velocity and acceleration constraints, in differentially flat space, gives us possibility to plan for position and yaw angle directly. Although the UAV is coupled with the manipulator, we rely on the controller to handle the manipulator motion as disturbance which allows us to plan the UAV trajectory in the differentially flat space.

Based on path and dynamic constraints, the output trajectory can be written as:
\begin{equation} \label{eqn:trajectory}
	\mathcal{T} = \left\{ \mathbf{w}^*(t) \mid \mathbf{w}^*(t) \in \mathbb{R}^{3(4+M)}, t \in (0,t_{end}) \right\},
\end{equation}
\noindent where $t$ denotes time. 

\subsection{End-effector pose compensation} \label{sec:end_effector_pose_comp}
\begin{figure}[t]
	\centering
	\subfloat[Difference between desired (red) and executed (blue) end-effector pose, without performing dynamic compensation. One observes that desired pose requires the UAV to horizontally without disturbing the hover condition.]{\includegraphics[trim={1cm 0.5cm 0.5cm 11cm},clip,width=\columnwidth]{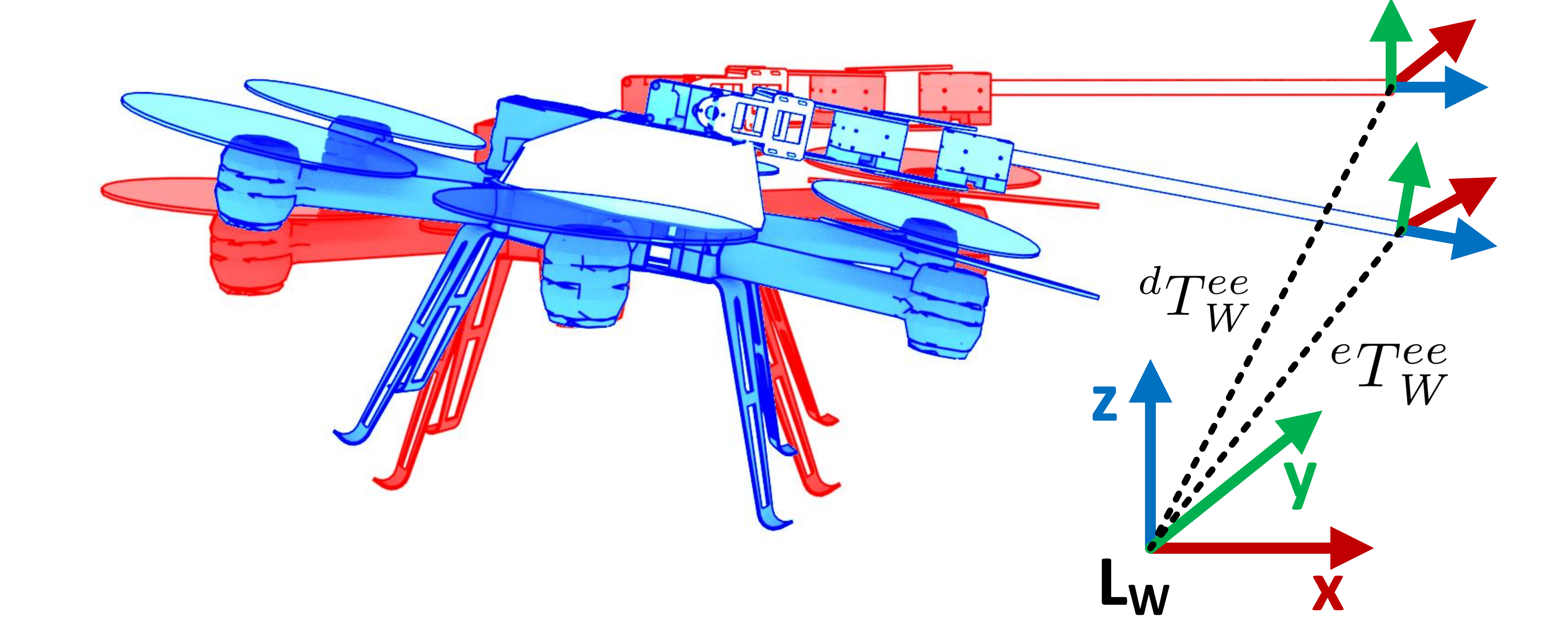}}
	\newline
	\subfloat[Difference between end-effector states with (yellow) and without compensation (blue). When we incorporate the full model of the UAV, which requires it to tilt in order to move horizontally, compensation is necessary to ensure the end-effector follows achieve the desired pose.]{\includegraphics[trim={1cm 0.5cm 0.5cm 11cm},clip,width=\columnwidth]{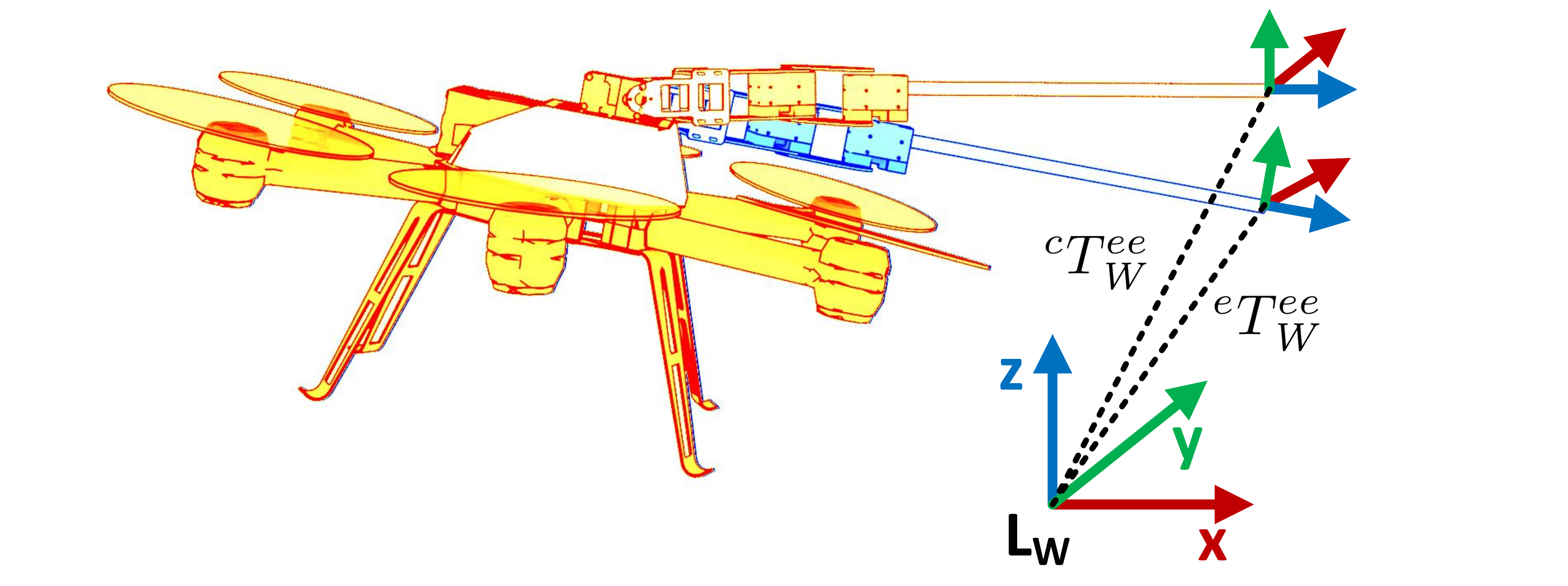}}
	\caption{An example of manipulator end-effector pose difference in planned and executed trajectory. Note that image depicts a single snapshot taken during trajectory execution. ${}^{d}T_{\mathit{W}}^{ee}$ is the desired end-effector pose, ${}^{e}T_{\mathit{W}}^{ee}$ is the end-effector pose for trajectory without compensation and ${}^{c}T_{\mathit{W}}^{ee}$ is the end-effector pose for trajectory with compensation.}
	\label{fig:compensation}
	\vspace{-0.45cm}
\end{figure}

As the aerial manipulator reaches certain waypoints, it will do so with some roll and pitch angles. In quasi-static motion, where dynamic constraints are small, the real and planned (control space) trajectory of the end-effector may not differ a lot. Depending on the task, this may be satisfactory behavior. However, even in the best of cases, this will force the use of conservative plans that execute for a long time and consume a lot of power since most of the power consumed is spent to keep the system in the air. In order to track the desired trajectory of the end-effector the pose errors induced by changes in the attitude angles are compensated using the manipulator's null space. This is done in two stages: first, the trajectory is simulated to obtain roll and pitch angles which are not taken into account while planning a trajectory; and second, exploiting the null space of the manipulator to find a solution as close as possible to the planned end-effector trajectory, depicted on Fig. \ref{fig:compensation}.

\subsubsection{Simulate trajectory execution}
The first step is to simulate the dynamics of the aerial manipulator by executing planned trajectory $\mathcal{T}$ described in equation \eqref{eqn:trajectory}. To simulate system realistically, the trajectory is sampled with the sampling frequency of the position controller and executed in simulation:
\begin{equation} \label{eqn:trajectory_sampled}
	\mathcal{T}_s = \left\{ \mathbf{w}^*(kT_s) \mid \mathbf{w}^*(kT_s) \in \mathbb{R}^{3(4+M)}, k \in (0,\dots,n_t)  \right\},
\end{equation}
\noindent where $T_s$ is sampling time of the trajectory and $k$ iterates through samples. During simulation, roll and pitch angles are collected and matched with every state in the sampled trajectory. By doing so we obtain the executed trajectory in terms of full generalized coordinates $\mathbf{w} = \begin{bmatrix} (\mathbf{q})^T & (\dot{\mathbf{q}})^T & (\ddot{\mathbf{q}})^T \end{bmatrix}^T \in \mathbb{R}^{3(6+M)}$:
\begin{equation} \label{eqn:trajectory_sampled_full}
	\mathcal{T}_{s,f} = \left\{ \mathbf{w}(kT_s) \mid \mathbf{w}(kT_s) \in \mathbb{R}^{3(6+M)}, k \in (0,\dots,n_t)  \right\}.
\end{equation}

\noindent Note that $\mathcal{T}_{s,f}$ still contains all elements of the planned trajectory from equation \eqref{eqn:trajectory_sampled} but is also enriched with roll and pitch angles obtained through simulation.

Having this information allows us to calculate the difference between the executed and the planned end-effector pose and compensate the error. It is worth mentioning that different dynamic models could be used, as long as they accurately depict the UAV body motion. For instance, one can use actual recorded flight data and use it to calculate the necessary compensation. This flight data can be recorded while flying in obstacle free environment and later utilized in constrained environment.

\subsubsection{Compensate the end-effector pose through manipulator null space}
Plugging \eqref{eqn:trajectory_sampled_full} into \eqref{eqn:kinematic_chain} we can obtain the desired pose of the end-effector in the world frame ${}^{d}{T}_\mathit{W}^{ee}$ for every point in the sampled trajectory. Rearranging \eqref{eqn:kinematic_chain} and plugging in \eqref{eqn:trajectory_sampled_full}, with information about the roll and the pitch angles from the simulated trajectory execution, we obtain the desired pose of the end-effector in $L_0$ coordinate system of the manipulator:
\begin{equation} \label{eqn:compensation}
	{}^{d}T_{L_0}^{ee} = (T_{\mathit{B}}^{L_0})^{-1} \cdot (T_{\mathit{W}}^{\mathit{B}})^{-1} \cdot {}^{d}{T}_\mathit{W}^{ee}.
\end{equation}
\noindent We can use the previous equation and the inverse kinematics of the manipulator to calculate the desired manipulator state vector ${}^{d}\mathbf{q}_M$. Fig. \ref{fig:compensation} illustrates the described algorithm. Note that we are compensating end-effector pose utilizing the manipulator null space and we change only the manipulator's portion of the planned trajectory with the new values. In case of having redundant DoFs, we choose the inverse kinematics solution that is closest to the manipulator configuration $\mathbf{q}_M$ in the previous step of trajectory. On the other hand, if there is no exact solution we use an approximated one instead.

\section{Experimental results} \label{sec:experimental}

\begin{table}[b]
\centering
\caption{DH parameters of the single-arm manipulator}
\label{table:dh_parameters}
\begin{tabular}{|c||c|c|c|c|}
\hline
 & $\theta$ & $d$ & $\alpha$ & $a$ \\
\hline
$1$ & $\pi/2$ & $0$ & $3\pi/2$ & $0.1365m$\\
\hline
$2$ & $0$ & $0$ & $0$ & $0.0725m$\\
\hline
$3$ & $3\pi/2$ & $0$ & $3\pi/2$ & $0m$\\
\hline
$4$ & $0$ & $0.4m$ & $0$ & $0$\\
\hline
\end{tabular}
\end{table}
For the experimental validation we use the \textit{AscTec NEO} hexacopter equipped with \textit{Intel NUC} onboard computer running ROS middleware. The low-level attitude controller, \textit{AscTec Trinity}, uses serial communication to exchange information with the onboard computer. The high-level model predictive position controller is implemented within the ROS middleware \cite{kamel2017}. \textit{Optitrack} motion capture system is used to obtain the position feedback of the UAV. 

To provide additional degrees of freedom, a 3 Degree-of-Freedom(DoF) single-arm manipulator is attached to the body of the UAV. The whole system is depicted in Fig. \ref{fig:neo_peg_in_hole}. The links of the manipulator are made from carbon fiber to reduce the weight while maintaining strength. The joints are actuated with \textit{Dynamixel XM430-W350R} servo motors, connected to the onboard computer through the USB interface. The attachment point of the manipulator on the UAV body is slightly above the UAV's center of gravity, to ensure stability during contact with the environment. The DH parameters necessary for the direct and inverse kinematics are provided in Table \ref{table:dh_parameters}. 

\begin{figure}[t]
\centering
\includegraphics[width=\columnwidth]{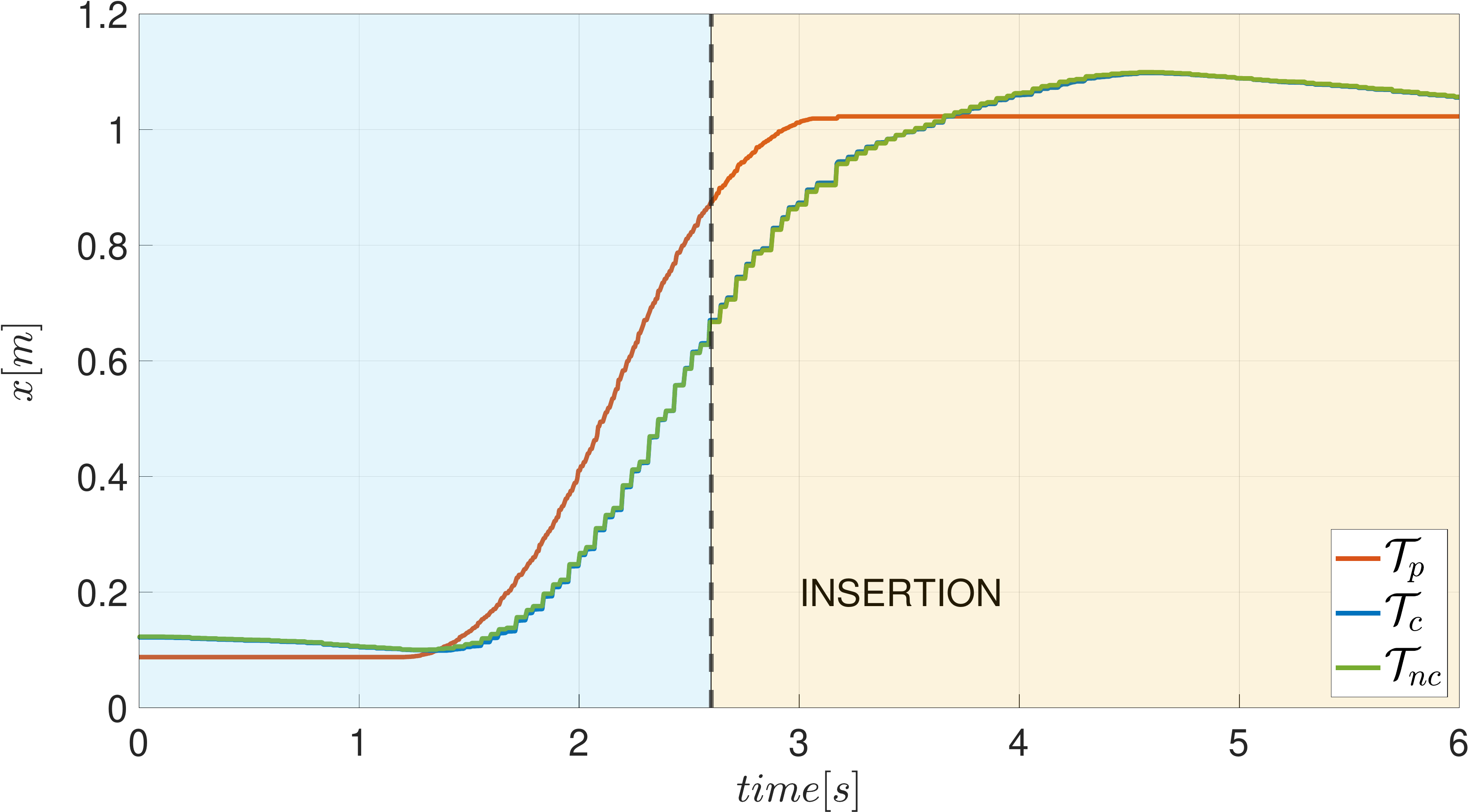}
\caption{The end-effector desired and executed pose in x-axis. There is no significant difference between compensated and non-compensated end-effector pose due to the small UAV pitch angle.}
\label{fig:exp_ravno_x}
\vspace{-0.45cm}
\end{figure}

\begin{figure}[t]
\centering
\includegraphics[width=\columnwidth]{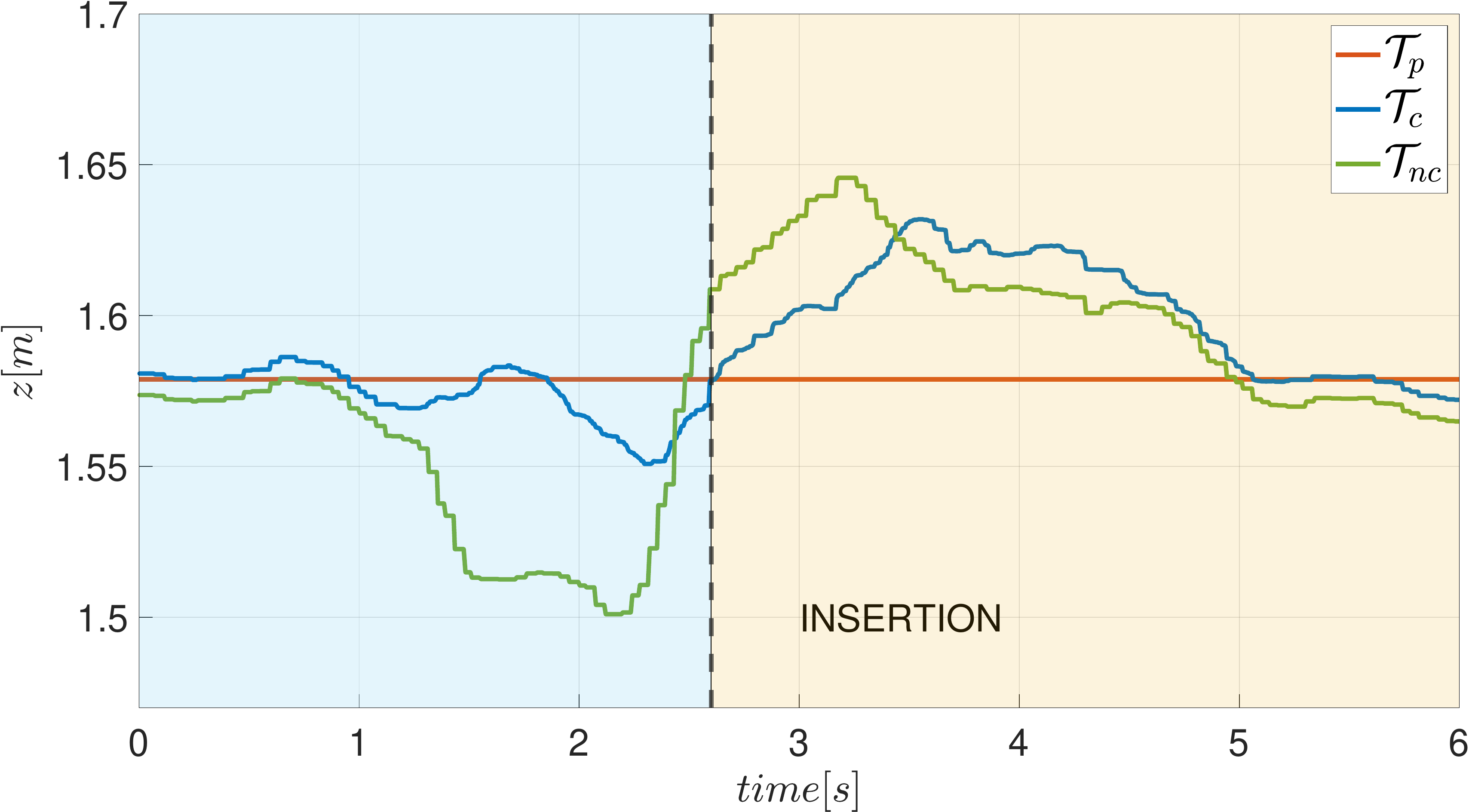}
\caption{Shows desired $\tau_p$ and executed end-effector pose in z-axis, with $\tau_c$ and without compensation $\tau_{nc}$, during insertion repeatability experiment. Similar results have been obtained during all five repetitions.}
\label{fig:exp_ravno_z}
\vspace{-0.45cm}
\end{figure}

To simulate the task constrained inspection scenario, we mounted a clear tube $7cm$ in diameter on a stand with a clear goal to perform a peg-in-hole insertion task shown in Fig. \ref{fig:neo_peg_in_hole}. We performed two experiments: the repeatability of the insertion in obstacle-free space; and insertion with the obstacle avoidance. 

Throughout the experiments we use the following notation: $\tau_p$ is the planned trajectory; $\tau_c$ is the executed trajectory with compensation of the end-effector pose included in the plan; and $\tau_{nc}$ is the same trajectory without compensation in end-effector pose. In other words we want to emphasise the importance of the compensation and visualize it during a single executed trajectory. 

\subsection{Insertion repeatability}
We tested the repeatability of the peg-in-hole insertion task in five sequential experiments. The UAV was repeatedly sent from starting point to the insertion point, which was in front of the UAV. Four out of five experiments were carried out successfully, while in the last one the insertion tube was missed. The Fig. \ref{fig:exp_ravno_x} shows the end-effector's desired $\tau_p$ and executed end-efector pose with $\tau_c$ and without $\tau_{nc}$ compensation. The difference between compensated and non-compensated trajectory is negligible in x-axis. This is the result of small pitch angle variations during UAV motion. But, even a small pitch angle offset can result in a large z-axis displacement, as shown in Fig. \ref{fig:exp_ravno_z}. This proves that the compensation is necessary for a successful completion of the task.

\subsection{Obstacle avoidance}
\begin{figure}[t]
\centering
\includegraphics[width=\columnwidth]{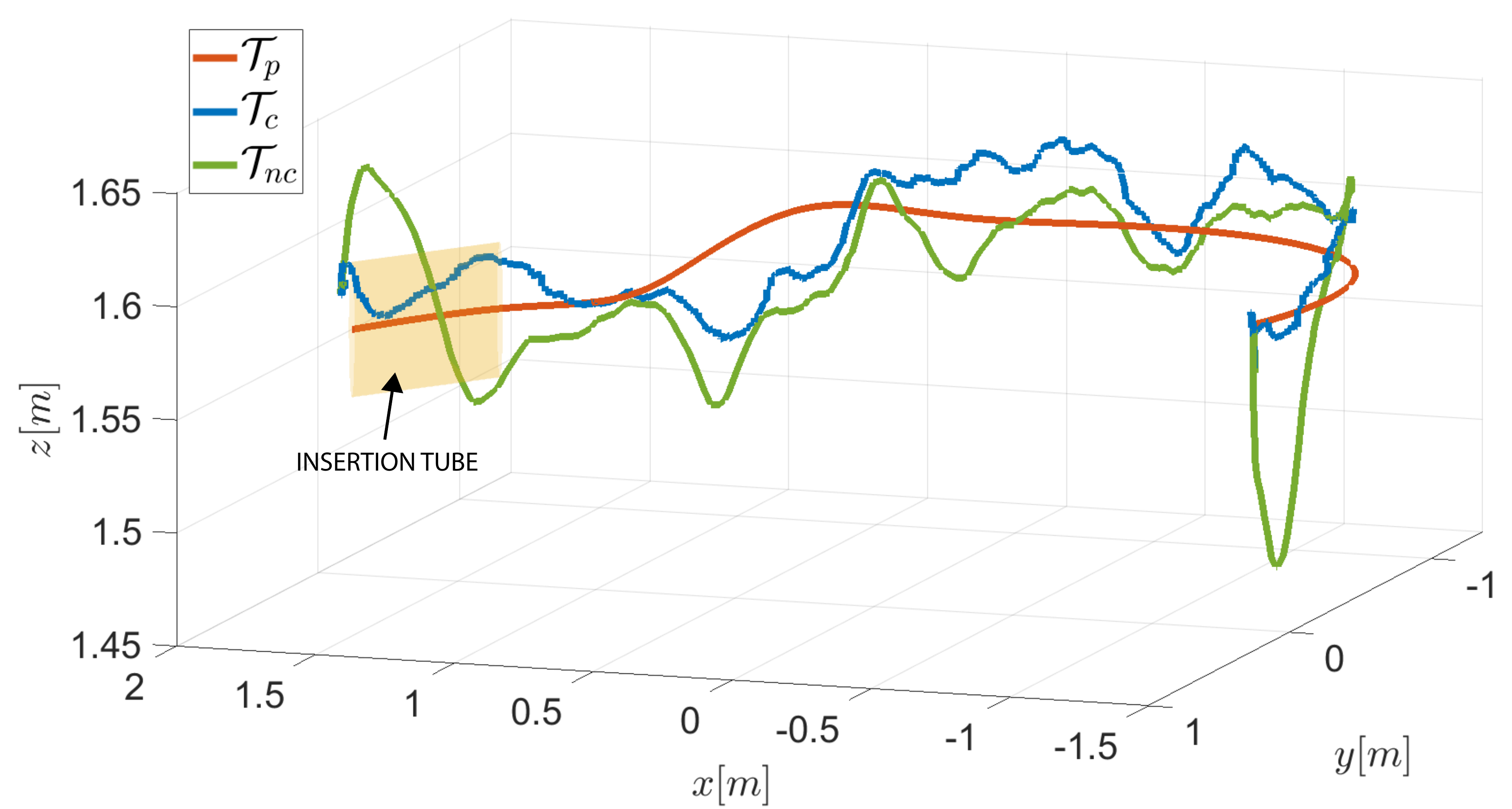}
\caption{The end-effector's poses during obstacle avoidance experiment. It can be observed that the experiment cannot be successfully conducted using an uncompensated trajectory. On the contrary, without trajectory compensation, the insertion point is missed.}
\label{fig:exp_labirint_3Dxyz}
\vspace{-0.45cm}
\end{figure}

\begin{figure}[t]
\centering
\includegraphics[width=\columnwidth]{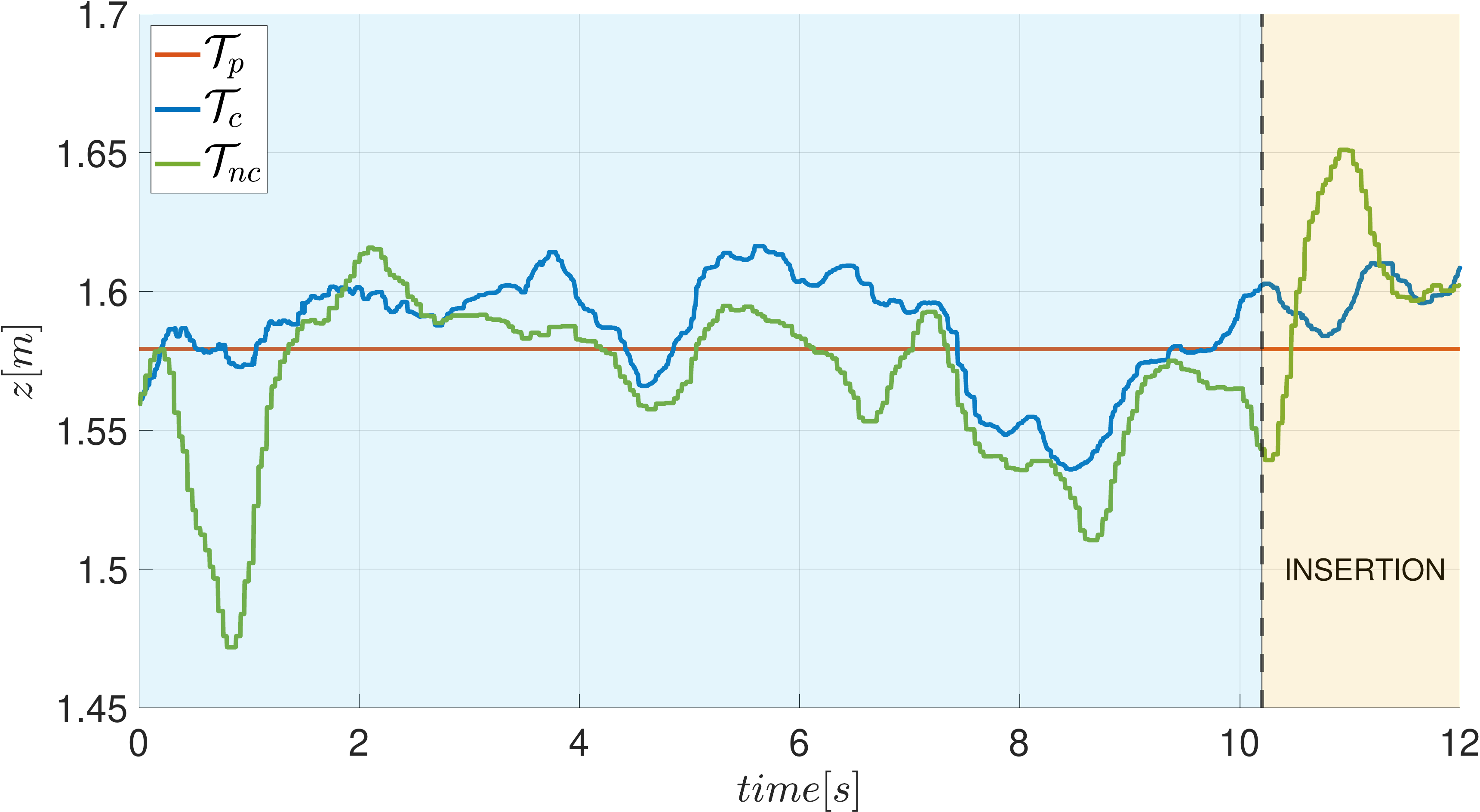}
\caption{Shows end-effector z-axis poses. It can be observed how compensated end-effector pose stays unchanged during UAV pitching, which is the contribution of the compensated trajectory.}
\label{fig:exp_labirint_z}
\end{figure}

In the second experiment, the compensation algorithm was verified in an obstructed environment. The insertion tube was placed inside a labyrinth with known insertion point. A compensated and an uncompensated trajectory were generated in order to compare task performance. The trajectories in 3D space together with insertion tube are shown in Fig. \ref{fig:exp_labirint_3Dxyz}. The close-up of the end-effector z-axis during experiment is shown in Fig. \ref{fig:exp_labirint_z}. It can be seen that the compensated end-effector pose stays unchanged during UAV movement. This effect is most emphasised at the beginning and the end of the trajectory, since these are the portions where the UAV starts accelerating. The video showing both experiments can be reached from \cite{videoIcra2020}.

\section{Conclusion} \label{sec:conclusion}
Within this paper, we have presented an aerial manipulator end-effector trajectory planning by exploiting the manipulator null space for compensating the end-effector state. The planning framework consists of three major phases: path planning; trajectory planning; and using null space for end-effector pose compensation. Using the RRT* in path planning allows for fast C-space exploration and obtaining feasible plans. The underactuated nature of the considered multirotor UAV introduces unknowns in full state of the aerial manipulator, namely in path planning phase the roll and pitch angles of the UAV are treated as unknowns. Planning a trajectory through TOPP-RA method introduces dynamical constraints, yielding a dynamically feasible trajectory. Simulating the obtained trajectory through the dynamical model of the system allows for obtaining the unknown roll and pitch angles. These, at this point known, dynamical effects introduce errors in the final end-effector planned pose which is compensated through redundancy of the manipulator by utilizing its null space. Although the joint angles of the manipulator deviate from the original plan, the end-effector pose is successfully compensated in the planning phase. This method is tested in experimental setup that verifies the proposed approach.



\section*{ACKNOWLEDGMENT}

This work has been supported in part by Croatian Science Foundation under the project Specularia UIP-2017-05-4042 \cite{SpeculariaWeb} and by European Commission Horizon 2020 Programme through project under G. A. number 810321, named Twinning coordination action for spreading excellence in Aerial Robotics - AeRoTwin \cite{AEROTWINweb}.


\bibliographystyle{ieeetr}
\bibliography{bibliography/bibliography}

\end{document}